\documentclass[letterpaper]{article} 
\usepackage{aaai25}  
\usepackage{times}  
\usepackage{helvet}  
\usepackage{courier}  
\usepackage[hyphens]{url}  
\usepackage{graphicx} 
\usepackage{natbib}  
\usepackage{caption} 
\usepackage{algorithm}
\usepackage{algorithmic}

\usepackage[utf8]{inputenc} 
\usepackage[T1]{fontenc}    
\usepackage{url}            
\usepackage{booktabs}       
\usepackage{amsfonts}       
\usepackage{nicefrac}       
\usepackage{microtype}      
\usepackage{xcolor}         

\usepackage{multirow}       
\usepackage{graphicx}
\usepackage{amsmath}
\usepackage{float}
\usepackage{epsfig}
\usepackage{subcaption}

\usepackage{newfloat}
\usepackage{listings}
\DeclareCaptionStyle{ruled}{labelfont=normalfont,labelsep=colon,strut=off} 
\floatstyle{ruled}
\newfloat{listing}{tb}{lst}{}
\floatname{listing}{Listing}
\title{{Transtreaming: Adaptive Delay-aware Transformer \\  for Real-time Streaming Perception}}
\author{
    Xiang Zhang$^1$,
    Yufei Cui$^2$,
    Chenchen Fu$^1$,
    Weiwei Wu$^1$,\\
    Zihao Wang$^1$,
    Yuyang Sun$^1$,
    Xue Liu$^2$,
}
\affiliations{
    $^1$ School of Computer Science and Engineering, Southeast University \\
    $^2$ School of Computer Science, McGill University
}

\usepackage{bibentry}

\begin{document}

\maketitle

\begin{abstract}
Real-time object detection is critical for the decision-making process for many real-world applications, such as collision avoidance and path planning in autonomous driving. This work presents an innovative real-time streaming perception method, Transtreaming, which addresses the challenge of real-time object detection with dynamic computational delay. The core innovation of Transtreaming lies in its adaptive delay-aware transformer, which can concurrently predict multiple future frames and select the output that best matches the real-world present time, compensating for any system-induced computation delays.

The proposed model outperforms the existing state-of-the-art methods, even in single-frame detection scenarios, by leveraging a transformer-based methodology. It demonstrates robust performance across a range of devices, from powerful V100 to modest 2080Ti, achieving the highest level of perceptual accuracy on all platforms. Unlike most state-of-the-art methods that struggle to complete computation within a single frame on less powerful devices, Transtreaming meets the stringent real-time processing requirements on all kinds of devices. The experimental results emphasize the system's adaptability and its potential to significantly improve the safety and reliability for many real-world systems, such as autonomous driving.
\end{abstract}

%
\begin{links}
    \link{\small{Code}}{https://anonymous.4open.science/r/Transtreaming-7333}
\end{links}

\section{Introduction} \label{sec:introduction}

The existing object detection methods have now achieved high performance within low latency. Most of these studies either focused on 1) improve the detection accuracy \cite{parmar_image_2018, liu2021swin}, or 2) make trade-off between the performance and the latency \cite{YOLO, yolox}. More recently, while real-world applications such as autonomous driving systems require real-time detection results for accurate decision makings, researchers began to pay much attention on real-time object detection. This area, also known as streaming perception, is gaining prominence for its potential to revolutionize the way we interact with and understand our environment in real time.   

For real-time object detection, researchers initially aimed at enhancing the speed of non-real-time detectors to meet stringent real-time requirements, such as achieving frame rates of $60$ frames per second (FPS). However, as pointed by \cite{StreamYOLO}, no matter how fast the detector is, in the real-world online scenario, when the detector handles the latest observation, the state of the world around the moving vehicle has already changed. Thus the detection result will always be at least one frame delay from the latest observation.


\begin{figure}
    \centering
    \includegraphics[width=0.8\columnwidth]{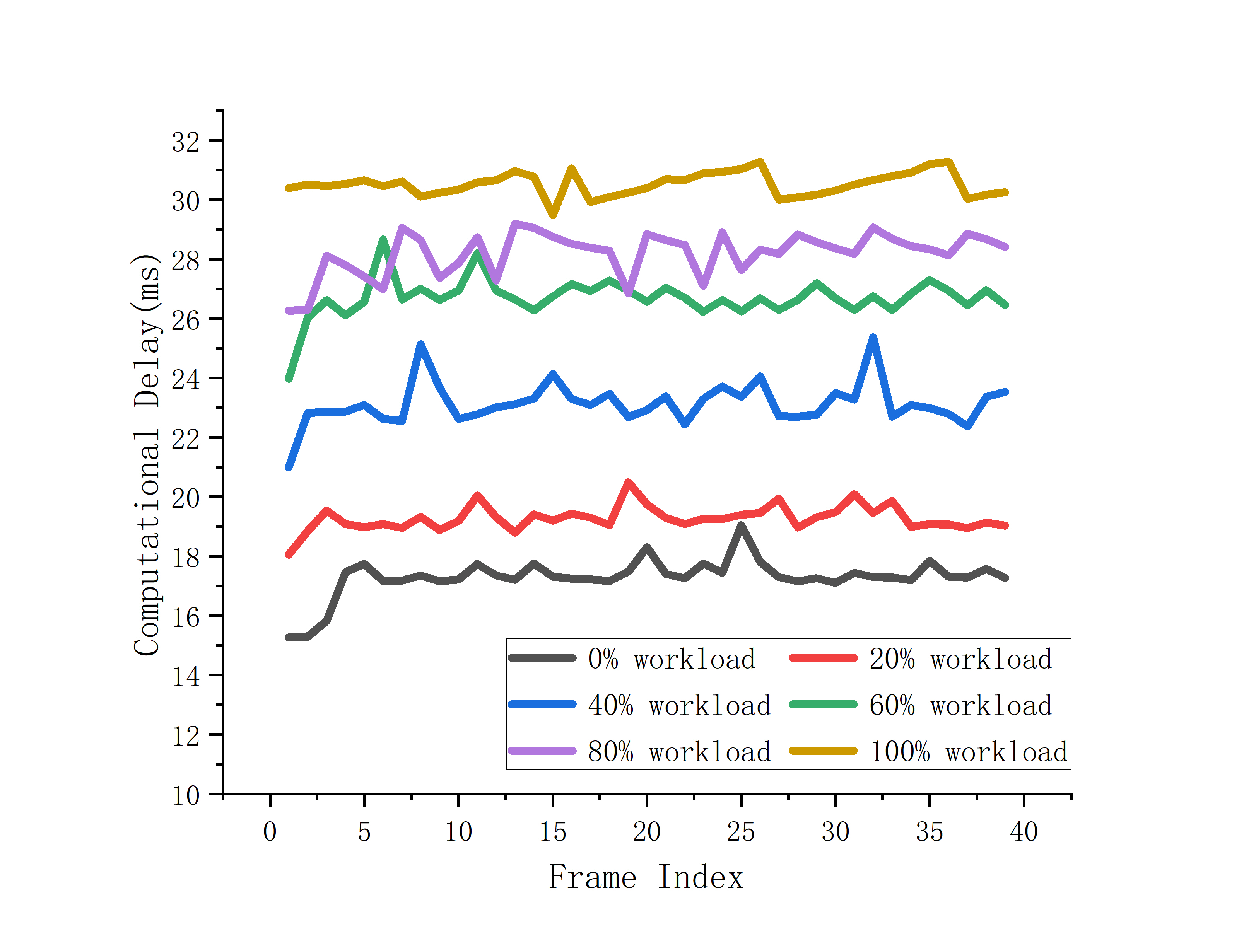}
    \vspace{-0.15in}
    \caption{\small Fluctuation of the computational delay for the same detector dealing with different frames with varying workloads in the server with NVIDIA GeForce RTX 4080.}
    \vspace{-0.2in}
    \label{fig:delay}
\end{figure}

To this end, a few recent studies \cite{StreamYOLO,Longshortnet,DAMO-StreamNet,DaDe,MTD} have improved the detector models to include predictive capabilities. These models are trained to anticipate the future positions of objects, for instance, one frame ahead, and make the detector finish the computation within one frame \cite{StreamYOLO}. For example, assuming a consistent video streaming speed, such as $30$FPS, this implies that the computation must be finished within $\frac{1}{30}$ second.
To emulate varying vehicular speeds, these models were trained across a spectrum of video streaming velocities. For example, if a vehicle is traveling at $30km/h$ (or $60km/h$), the video streaming is adjusted to match the speed, set at $1\times$ (or $2\times$) the normal rate, respectively. 
Despite these advancements, the above literature presents several limitations:
\begin{enumerate}
    \item The detectors are required to complete computations within a constrained timeframe, which requires that most of the models (especially those with high performance) must run in devices with powerful GPUs like V100 \cite{StreamYOLO, Longshortnet, DAMO-StreamNet}. This requirement restricts the application of real-time detection systems to devices with substantial computational resources. 
    \item Existing models are typically restricted to predicting a fixed single frame in the future, such as one or two frames ahead. However,  we observe that the computational delays can fluctuate significantly over time due to varying workloads. As shown in Figure \ref{fig:delay}, the same detector has different computational delays in dealing with different frames, ranging from $15ms$ to $31ms$ due to various workloads of the system. This variability implies that models may not be able to meet fixed real-time constraints during periods of burst workload, potentially leading to significant failures in accurate real-time detection.
\end{enumerate}

To address the above limitations, this work presents a novel real-time streaming perception method, Transtreaming, which addresses the challenge of real-time object detection with dynamic computational delay. The contributions of this work are summarized as follows.

\begin{itemize}
    \item This work facilitates adaptive delay-aware streaming perception by integrating runtime information inside detection models, concurrently predicting {\bf multiple frames} and selecting the most appropriate output aligned with the real-world present time.
    \item Even in the realm of {\bf single-frame} real-time detection, the proposed approach, which leverages a transformer-based methodology, outperforms all the existing state-of-the-art (SOTA) techniques.
    \item The proposed model demonstrates robust performance across a spectrum of devices, ranging from the high-performance V100 to the modest 2080Ti. It achieves the highest level of perceptual accuracy on all platforms, whereas most SOTA methods struggle to complete computation within a single frame when applied on less powerful devices, thereby failing to meet the stringent real-time processing requirement.
\end{itemize}


\section{Related Work} \label{sec:related_work}

{\bf Image Object Detection:}
Deep learning's evolution has dramatically transformed object detection, with CNNs outpacing traditional methods. Image object detection methods are primarily divided into two-stage, exemplified by R-CNN \cite{girshick_rich_2014}, which enhanced accuracy through region proposals combined with CNN features, and one-stage, such as SSD \cite{SSD} and YOLO \cite{YOLO}. Improvements in the two-stage approach, seen in Fast R-CNN \cite{FastRCNN} and Faster R-CNN \cite{FasterRCNN}, streamlined the process by merging region proposal networks with CNN architectures, reducing latency. Yet, these techniques still face delays due to proposal refinement. One-stage detectors offer a balance of speed and precision for real-time tasks but lack the temporal context necessary for streaming detection, as they focus solely on the current frame.


{\bf Video Object Detection:}
Video Object Detection (VOD) aims to optimize detection on poor-quality video frames, such as those with motion blur or defocus. It is divided into four strategies: 1) tracking-based methods \cite{D&T}: enhance detection by learning feature similarities across frames; 2) optical-flow-based methods \cite{FGFA, MANet, THP, DFF}: utilize flow information to enhance key frames; 3) attention-based methods \cite{RDN, SELSA, LLTR, HVRNet}: apply attention mechanisms to relate abstract features for improved detection; and 4) other methods: aggregate features through networks like DCN \cite{DCN}, RNN \cite{RNN}, and LSTM \cite{LSTM}.
VOD generally focuses on the offline setting, often overlooking the runtime delay of detection algorithms. In contrast, this work takes into account both runtime delay and real-time requirements to ensure high-performance detection.

{\bf Streaming Perception:}
Streaming Perception tackles the drift in real-time detection due to computational latency by predicting entity locations after computational delays using temporal information from historical results \cite{Towards_Streaming_Perception}. The sAP metric was introduced to evaluate object detection algorithms in streaming scenarios, factoring in both latency and accuracy. Subsequent research introduced several models aimed at forecasting object locations. For example, StreamYOLO \cite{StreamYOLO} used a dual-flow perception module for next-frame prediction, combining both previous and current frames' features. Dade \cite{DaDe} and MTD \cite{MTD} introduced mechanisms to dynamically select features from past or future timestamps, taking the runtime delay of the algorithm into consideration. DAMO-StreamNet \cite{DAMO-StreamNet} and Longshortnet \cite{Longshortnet} used dual-path architectures to capture long-term motion and calibrate with short-term semantics. They extended dynamic flow path from 1 past frame to 3 frames, successfully capturing the long-term motion of moving objects, and thus achieved the state-of-the-art performance in streaming perception. 
These existing studies, however, only do the detection in the single next frame and thus force the detector must finish the computation within 1 frame all the time, and this is hard to achieve when the computing resources limited or the system workload is high. In this work, we propose a temporal attention module capable of predicting multiple future outcomes from any past time series, selecting the outcome that best aligns with the real-world present time, thereby adapting to various computational delays.


\begin{figure*}[t!]
    \centering
    \begin{subfigure}[t]{0.42\textwidth}
		\centering
		\includegraphics[width=1.0\textwidth]{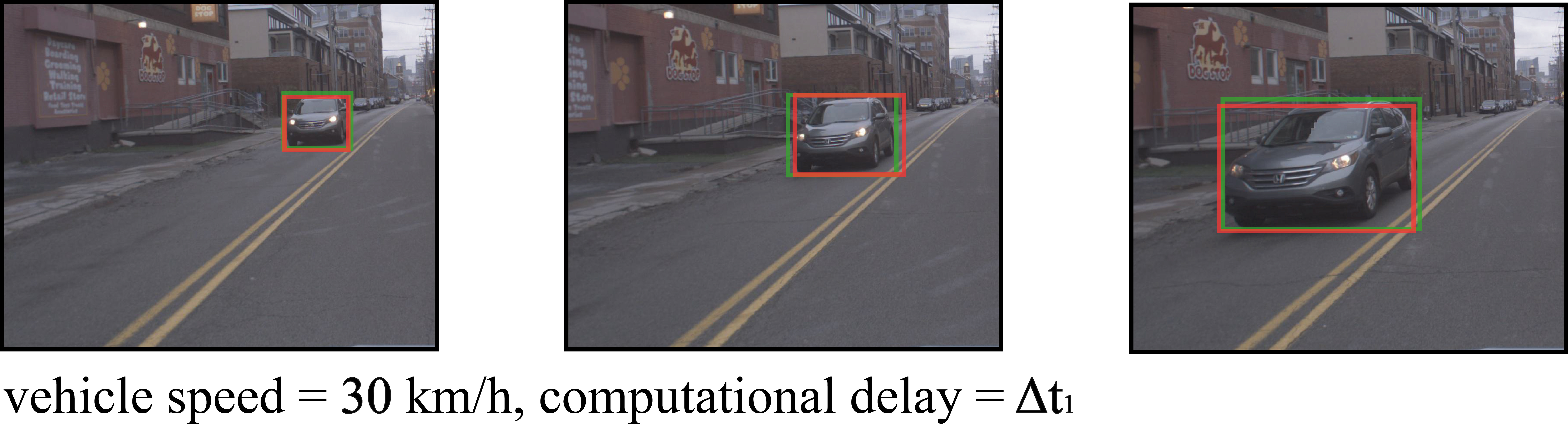}
		\includegraphics[width=1.0\textwidth]{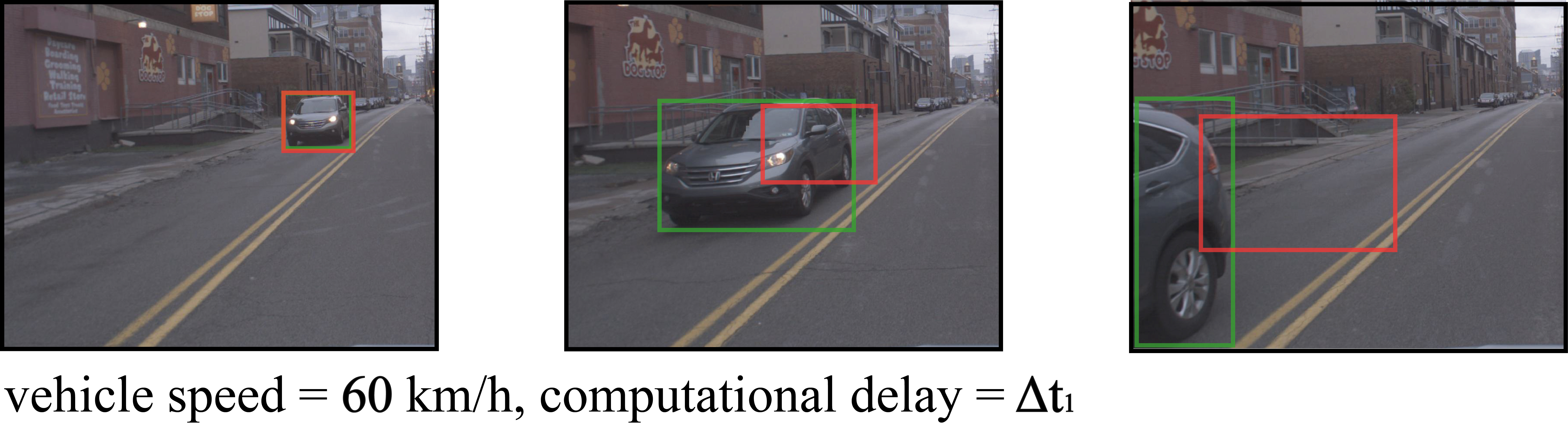}
		\caption{Given fixed $\Delta t_1$, varying speed reduces the accuracy}
	\end{subfigure}
    \hspace{3em}
    \begin{subfigure}[t]{0.42\textwidth}
    	\centering
    	\includegraphics[width=1.0\textwidth]{images/figure1_1.jpg}
    	\includegraphics[width=1.0\textwidth]{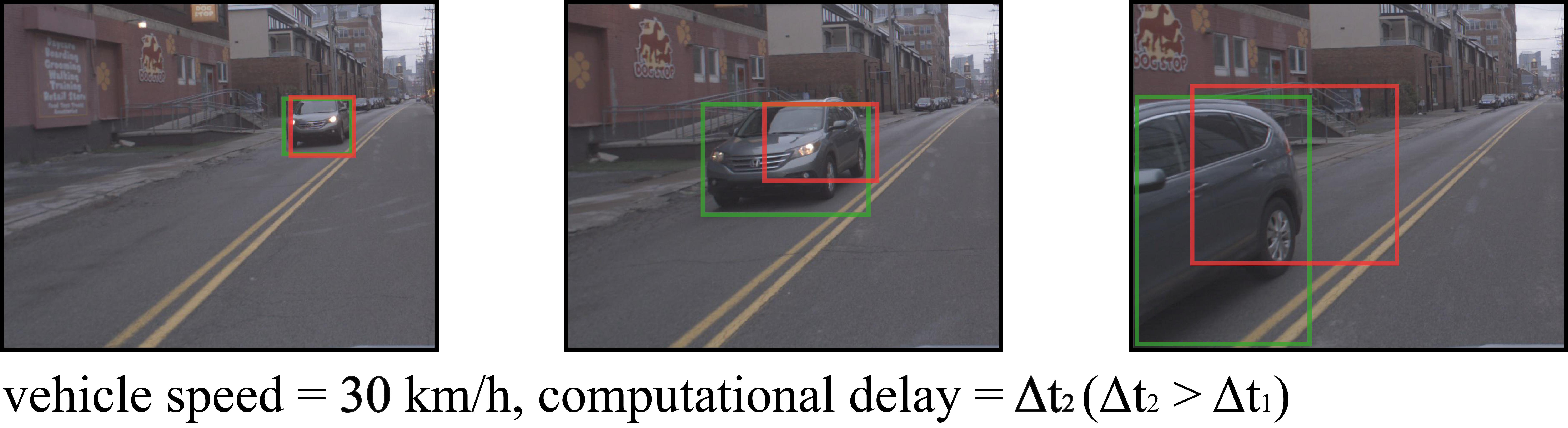}
    	\caption{\small Given fixed speed, dynamic delay reduces the accuracy}
    \end{subfigure}
    \caption{\small Observation example: when a detector is trained with fixed vehicle speed ($30km/h$) and fixed computational delay ($\Delta t_1$), the detection is always accurate when both of the speed and delay remain the same. But the detection accuracy will significantly decrease when either (a) the vehicle speed changes (from $30km/h$ to $60km/h$) or (b) the computational delay varies (from $\Delta t_1$ to $\Delta t_2$).}
    \vspace{-0.1in}
    \label{fig:motivation} 
\end{figure*}

\section{Methodology}
\label{sec:methodology}

\subsection{Observation and Motivation}

The motivation for our proposed scheme is rooted in the limitations observed in current streaming perception studies \cite{StreamYOLO, Longshortnet, DAMO-StreamNet}, particularly in handling the variable velocities of vehicles and the dynamic computational delays inherent to real-time systems.

The above SOTA methods in streaming perception have noticed that the different velocities of vehicles significantly affect the real-time detection. If the detector is only trained at a fixed frame rate (e.g., $60$FPS), it can only be equipped to simulate vehicles moving at a corresponding constant speed (e.g., $30km/h$). As depicted in Figure \ref{fig:motivation}.(a), when a vehicle's speed increases to $60km/s$, the detector fails to accurately detect it.
Even though mix-velocity training were applied in these work, they need to pre-set the velocity \cite{StreamYOLO}, i.e. pre-determine whether they do the detection for frame $t+1$ for slow or $t+3$ for fast velocities based on prior frames. These approaches are inherently restrictive because they did not allow dynamic velocity of vehicles.

Furthermore, as shown in Figure \ref{fig:delay}, random computational delay can be involved because of dynamic workloads in time. And thus even the vehicle drives in the same speed, different computational delay will result in failure in accurate detection, as shown in Figure \ref{fig:motivation}.(b). Notably, both varying vehicle velocities and computational delays introduce a similar element of randomness to the detection process, complicating the prediction task.

To address these challenges, we introduce Transtreaming, a novel approach that captures the inherent randomness in both vehicle velocity and computational delay. Our solution is designed to provide real-time detection forecasts across a range of temporal conditions, adapting dynamically to the operational environment. Specifically, Transtreaming is tailored to accommodate diverse runtime scenarios, thereby enhancing the safety and reliability of streaming perception systems in various operational contexts.

\begin{figure*}
    \centering
    \includegraphics[width=0.8\textwidth]{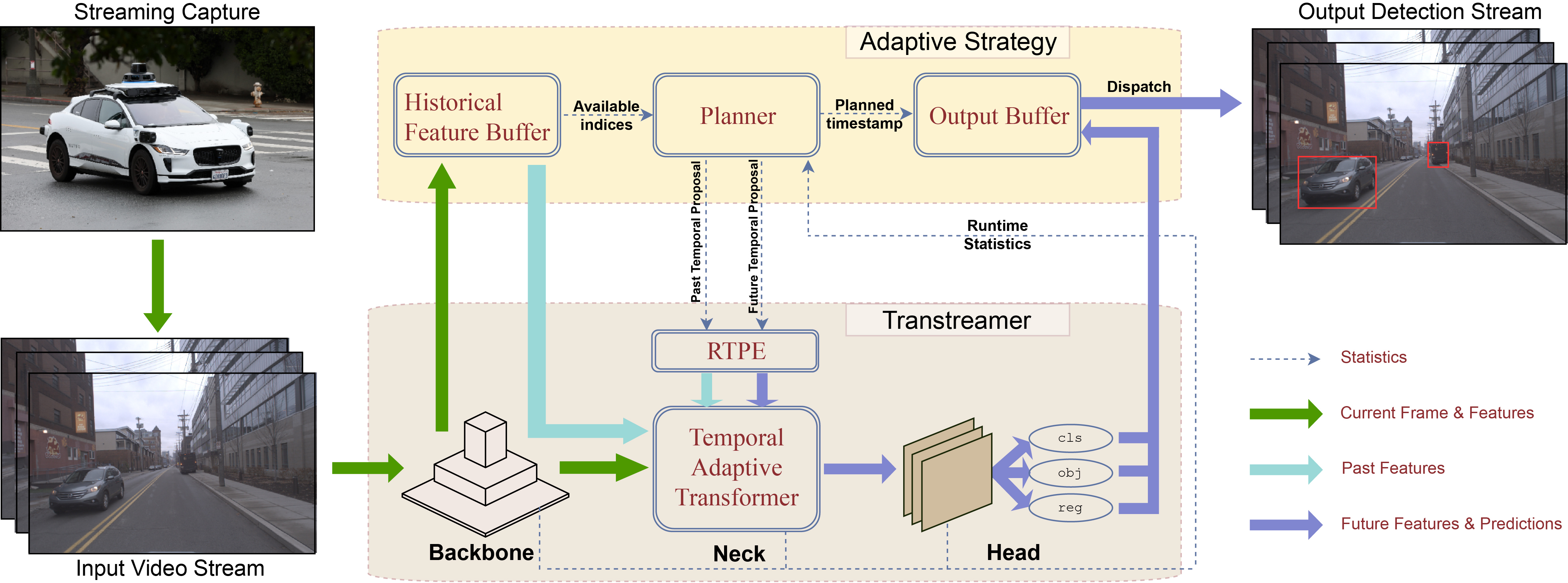}
    \caption{\small Overall architecture of Transtreaming. Transtreaming composes of a detection model \rm Transtreamer and strategy algorithm \rm Adaptive Strategy. The detection process is mainly done by Transtreamer in the lower part of the figure. Adaptive Strategy provides support by producing temporal proposals that encoded by RTPE (Relative Temporal Positional Embedding) and by using 2 buffers to store historical features and dispatch detection results.}
    \vspace{-0.1in}
    \label{fig:architecture}
\end{figure*}

\subsection{Problem Formation}
Given a RGB video sequence $\{\mathbf{I}_i\} \in \mathbb{R}^{L \times 3 \times H \times W}$ and the ground truth object bounding boxes $\{\mathbf{O}_i\} \in \mathbb{R}^{L \times Objects \times 5}$, each frame $\mathbf{I}_i$ and bounding box $\mathbf{O}_i$ at index $i$ is firstly associated with a timestamp $t=\frac{i}{k}$ with frame rate $k$ ($k=30$, typically) FPS under streaming perception settings. The task of streaming perception then runs an online procedure that delivers the frames to the detector at their corresponding timestamp $t$. Then the detector's predictions $\{\hat{O}_{t'}\}$ is collected at timestamp $t'$. Note that the detector has computational delay $\Delta t^I$ and start-up delay $\Delta t^S$ (delay caused by not inferring $\mathbf{I}_i$ at exactly $t$ because of delay from the previous loop). Consequently $t'=t+\Delta t^I+\Delta t^S$, making detector output timestamps $t'$ misaligned with $t$. As a final metric, streaming perception uses Mean Average Precision but pairs prediction and ground truth differently. Each prediction $\hat{\mathbf{O}}_{t'}$ is paired with the ground truths after its output timestamp $t'_1$ before the subsequent prediction's output timestamp $t'_2$, which is $\{\mathbf{O}_i\}, i \in [\lceil t'_1 \times k \rceil, \lfloor t'_2 \times k \rfloor]$. Unpaired ground truths are considered detection misses. Therefore, the detector is expected to generate bounding box prediction via forecasting into the future, in order to align $\hat{\mathbf{O}}_{t'}$ with $\mathbf{O}_{\lceil t' \times k \rceil}$. Every prediction of the detector should manage to compensate the object displacement error cause by computational delay $\Delta t^I$, and subsequent start-up delay $\Delta t^S$.
Without loss of generality, we simplify notations by using relative indices $\Delta i$ against current frame index. In this way, current frames is denoted as $\mathbf{I}_0$, past frames are denoted as $\mathbf{I}_{-1}, \mathbf{I}_{-2}, \dots$, while future objects are denoted as $\mathbf{O}_{1}, \mathbf{O}_{2}, \dots$.

The overall architecture of Transtreaming is shown in Figure \ref{fig:architecture}. We split the design of Transtreaming to a detection model named Transtreamer (TS) and a scheduling algorithm named Adaptive Strategy (AS). At every timestamp, given previous input frames $\{\mathbf{I}_{i}\}, i \in [0, \lfloor t * k \rfloor]$, Transtreaming utilizes AS to estimate runtime statistics $\Delta \hat{t}^I$ and schedule the execution of TS. To provide the detection model with temporal cues, AS generates 2 temporal proposals $\{\mathbf{P^P}, \mathbf{P^F}\}, \mathbf{P^P} \subseteq \mathbb{N}^-, \mathbf{P^F} \subseteq \mathbb{N}^+$ from estimated $\Delta \hat{t}^I$ and observed start-up delay $\Delta t^S$. Proposal $\mathbf{P^P}$ indicates the indices of available past frames' features, while $\mathbf{P^F}$ indicates the temporal indices of forecasting targets. AS also uses buffering techniques to store previously computed image features $\{\mathbf{F}_i\}, i \in \mathbf{P^P}$, in order to accelerate model inference speed. Furthermore, the multi-frame output of TS $\{\hat{\mathbf{O}}_j\}, j \in \mathbf{P^F}$ is stored in output buffer and dispatched at appropriate timestamp. This approach avoids producing $\hat{\mathbf{O}}_j$ before its actual timestamp. Inside the detection model, TS encodes the temporal proposals $\{\mathbf{P^P}, \mathbf{P^F}\}$ as Relative Temporal Positional Embedding (RTPE) and merge features buffers with current frame's features. TS then uses a temporal transformer module to query the features of future objects from current and past features with RTPE. Additionally, Transtreamer is capable of forecasting object on multiple future timestamps with dynamic length (i.e., $\mathbf{P^F}$ can have any number of elements), thus achieving flexibility in diverse runtime environments. The details of Transtreaming's components are elaborated in the following subsections.

\subsection{Transtreamer}
Following previous work's \cite{DAMO-StreamNet} design of detection models, we split TS into three modules: backbone for feature extraction, neck for temporal forecasting and head for bounding box decoding. To develop a neck module that maximally enhance temporal reasoning in TS, we employ the transformer paradigm to establish relationship between past features and future features. Transformers can effectively utilize the temporal proposals provided by AS. To evaluate against the current SOTA methods, we leverage the same backbone (DRFPN) and head (TALHead) module as in \cite{DAMO-StreamNet} and \cite{StreamYOLO}. Assume the temporal proposals $\{\mathbf{P^P}, \mathbf{P^F}\}$, and buffered features $\{\mathbf{F}_i\}, i\in \mathbf{P^P}$ is known beforehand. The features of current frame $\mathbf{I}_0$ is extracted by the backbone module $\mathbf{W^B}(\cdot)$. The resulting features $\mathbf{F}_0$ is then concatenated to the buffered features in a chronological order. Given the concatenated features and $\mathbf{P^P}, \mathbf{P^F}$, TS encodes $\mathbf{P^P}$ and $\mathbf{P^F}$ as Relative Temporal Positional Embedding (RTPE, denoted as $\rm RTPE$). $\rm RTPE$ carries temporal information for the succeeding Temporal Adaptive Transformer (TAT, denoted as $\mathrm{W^T}(\cdot)$). TAT makes use of cross attention modules that decodes future objects' features $\{\hat{\mathbf{F}}_{j}\}, j \in \mathbf{TF}$ from past frames' features. Finally, $\{\hat{\mathbf{F}}_{j}\}$ is fed into the head module $\mathrm{W^H}(\cdot)$ to generate the estimated bounding boxes for future objects $\{\hat{\mathbf{O}}_{j}\}$. 

\begin{figure*}
    \centering
    \includegraphics[width=0.8\textwidth]{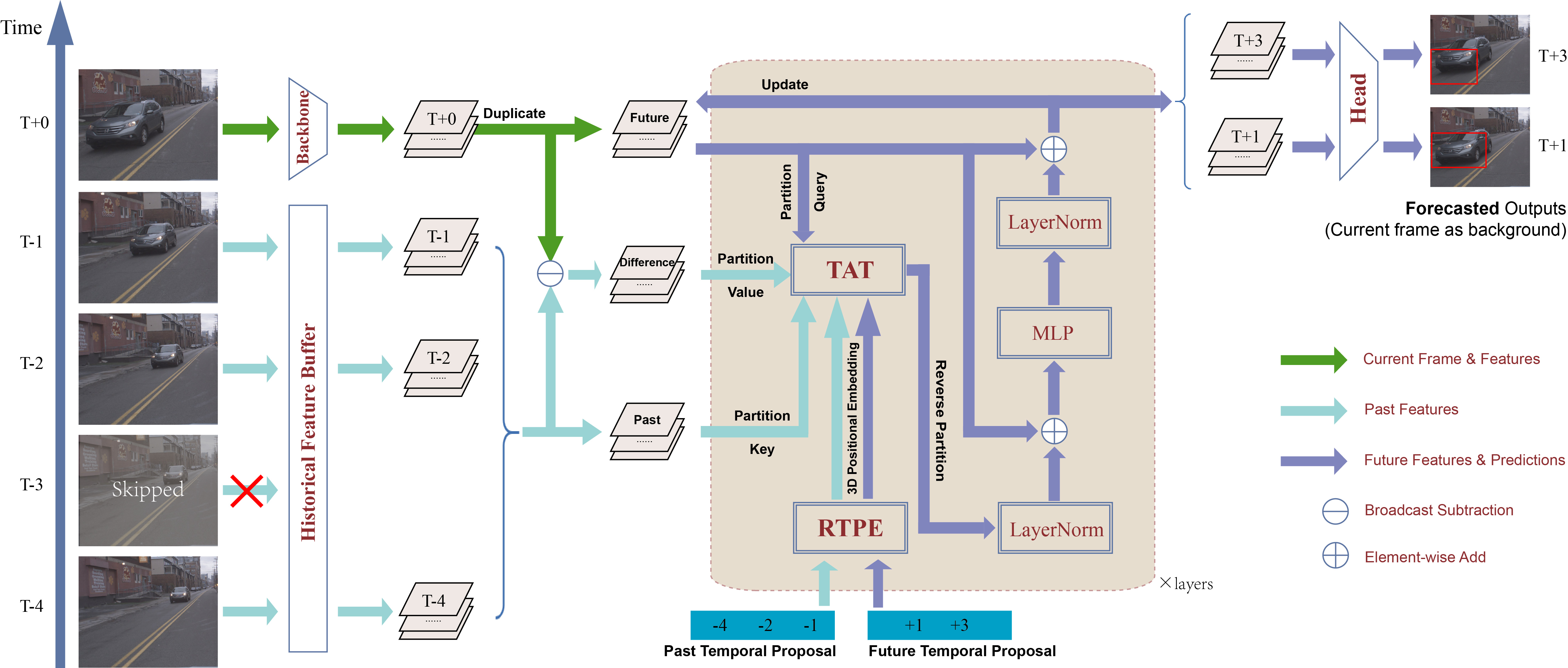}
    \caption{\small Detailed architecture of the detection model: Transtreamer. The example illustrates the situation that $T-3$ frame is skipped due to computational delay, and Adaptive strategy asks Transtreamer to produce the detection result of $T+1$, $T+3$ from $T+0$ image and $T-1$, $T-2$, $T-4$ buffered features.}
    \vspace{-0.1in}
    \label{fig:model}
\end{figure*}

Formally, Transtreamer can be represented as
\begin{equation} \label{eq:model}
\begin{split}
    \{\hat{\mathbf{O}_j}\} & = \rm Transtreamer(\mathbf{I}_0, \{\mathbf{F}_i\}, \rm RTPE) \\
    & = \mathrm{W^H}(\mathrm{W^T}(\rm Concatenate(\mathrm{\mathbf{F}^B}(\{\mathbf{I}_0\}), \{\mathbf{F}_i\}), \rm RTPE), \\
\end{split}
\end{equation}
where $i \in \mathbf{P^P}$, $j \in \mathbf{P^F}$. The whole architecture of TS is shown in Figure \ref{fig:model}.

\vspace{3pt}
{\bf 3D Window Partitioning:}
The extracted past and current features $\{\mathbf{F}_i\}, i \in \mathbf{P}^P + \{0\}$ consist of 2 spatial dimensions (height $H$ and width $W$), 1 temporal dimension (time $T$) and a feature dimension (channel $C$). Common tokenizing approach flattens the input into a $THW \times C$ token sequence. But this approach is computational expensive with a computing complexity of $O(T^2H^2W^2C)$. To reduce the complexity of attention computation, we leverage the windowed attention to break input features to 3d spatial-temporal windows $(win^T, win^H, win^W)$. We assume the objects' trajectories can be captured within the window size, so shrinking the transformer's perception range can have negligible effect. In practice, we apply window partitioning before every layer of TAT and reverse window partitioning after that.

{\bf Relative Temporal Positional Embedding:}
The order of input sequence does not affect the computation of attention weight in attention modules. Therefore, the spatial and temporal locality of 3d features cannot be preserved. To remedy this problem, we employ positional embedding to integrate locality information into aforementioned Temporal Transformer by directly adding to attention weights. Since the concatenated features $\{F_i\}$ involves relative spatial and temporal position, we extend 2d learnable relative positional embedding used in \cite{liu2021swin} to 3d, adding an additional temporal dimension. Different from spatial dimensions, our RTPE only offers embedding for positive temporal difference (i.e. future tokens attend to past tokens, but not vice versa). Formally, the embedding created by RTPE is denoted as $E_{dt, dh, dw}$. $E_{dt, dh, dw}$ is generated by MLP projections from cuboid 3d coordinates $(t, h, w), t \in [0, max(\mathbf{P^F})-min(\mathbf{P^F})], h \in [-win^H, win^H], w \in [-win^W, win^W]$ to ensure its continuity in 3 dimensions. Here $max(\mathbf{P^F})-min(\mathbf{P^F})$ is the maximum temporal difference given $\mathbf{P^P}$ and $\mathbf{P^F}$. For each query token $Q_{t_1,h_1,w_1}$, its temporal embedding value against key token $K_{t_2,h_2,w_2}$ is $E_{t_1-t_2,h_1-h_2,w_1-w_2}$, effectively reflecting their relative positions.

{\bf Temporal Adaptive Transformer:}
The neck of Transtreamer should model the temporal relationship between past frames' features $\{\mathbf{F}_i\}$ and future objects' features $\{\mathbf{F}_{j}\}$, so directly using convolution or linear layers as \cite{DAMO-StreamNet, Longshortnet} will result in a fix-lengthed $\mathbf{P^P}$ and $\mathbf{P^F}$ which goes against our original intention. Inspired from the decoder layers of the transformer, we introduce a novel temporal cross-attention module TAT, which encodes future objects' features as queries while past frames' features as keys and values. Since attention modules often utilize residual connection to help mitigate the vanishing gradient problem, we argue that such approach is still valid in this situation. This is due to the fact that the features of current frame $\mathbf{F}_{0}$ is a reasonable initial guess of the feature of future frames $\{\mathbf{F}_{j}\}$. Specifically, we firstly replicate current features $\mathbf{F}_{0}$ along the temporal dimension by $|\mathbf{P^F}|$ as the initial queries. Past frames' features $\{\mathbf{F}_i\}$ are used as keys while the difference of $\mathbf{F}_{0}$ and $\{\mathbf{F}_i\}$ are used as values. Secondly, in every layer of TAT, 3D Window Partitioning is utilized to create corresponding input tokens $\mathbf{F}^Q$, $\mathbf{F}^K$ and $\mathbf{F}^V$. We adopt a cross attention with $\rm RTPE$ as gain to attention weight, serving as a source of temporal information. The tokens are transformed back to their original form by reverse window partitioning, followed by normalization layers and MLP layers as in original transformer architecture. Finally, we update the future queries with layer output and repeat the actions within every layer of TAT.

\subsection{Adaptive Strategy}
Adaptive Strategy, which consists of a planner module and two buffers, defines the way Transtreamer interacts with streaming perception environments.

{\bf Planner:}
The main function of AS is planning, in other words, is generating a proper $\mathbf{P^P}$ and $\mathbf{P^F}$ that maximize the detection model's performance in Streaming Perception. During streaming inference, the past loops' inference time of Transtreamer's backbone, neck and head module, as well as other processing delay is firstly recorded as $\{\Delta t_i^B, \Delta t_i^N, \Delta t_i^H, \Delta t_i^O\}, i \in \mathbb{N}^-$, respectively. As a reasonable guess, we estimate these delay in current loop $\{\Delta \hat{t}_0^B, \Delta \hat{t}_0^N, \Delta \hat{t}_0^H, \Delta \hat{t}_0^O\}$ with exponential moving average (EMA) of $\Delta t_i$ with a decay of 0.5. Additionally, the start-up delay can be directly measured as $\Delta t_0^S$. To fully utilize past buffers and to give a rational estimate of future timestamps, We select at most $M^P$ latest available buffers' index as $\mathbf{P_0^P}$. For $\mathbf{P_0^F}$, we produce at most $M^F$ indices, with each prediction evenly spaced on temporal axis. Finally, both $\mathbf{P^P}$ and $\mathbf{P^F}$ are clipped within the range of $[\Delta t_{min} \times k, \Delta t_{max} \times k]$ and remove duplicates to keep $\mathbf{P}$ in reasonable range and avoid redundant computation. This operation is denoted as $\rm Clip(\cdot)$. The value of $\Delta t_{min}$ and $\Delta t_{max}$ are empirically assigned as $-29$ and $+19$, respectively. This solution manages to maintain the balance between model latency and forecasting accuracy with respect to high and fluctuating delays.

{\bf Historical Feature Buffer:}
Though the features of past frames $\{\mathbf{F}_{i}\}, i \in \mathbf{P^P}$ can be acquired by computing $\{\mathbf{W^B}(\mathbf{I}_{i})\}$, such method is not applicable in streaming evaluations, where the computation load clearly affects the final performance. Therefore, we cache historical frame's feature to avoid re-computation. We also set a limit to the length of the buffer and choose to discard the earliest buffered features after the buffer is full. Since early frames have a negligible influence on the current frame, the limitation ease the computation burden of TAT.

{\bf Output Buffer:}
Although multiple future object predictions $\{\hat{\mathbf{O}}_j\}, j \in \mathbf{P^F}$ are produced by Transtreamer, they should not be emitted simply in sequential order. Because it may produce outdated result when $\Delta \hat{t}^I$ is underestimated. Therefore, we apply another buffer to the output of Transtreamer. At every timestamp, most temporally adjacent predictions in the buffer are dispatched, in order to effectively utilize all the predictions at appropriate timestamps.

\subsection{Training and Inference}
We replicate the training scheme used by baseline methods \cite{DAMO-StreamNet, Longshortnet}, with the exception of Asymmetric Knowledge Distillation proposed by \cite{DAMO-StreamNet}. We believe this contribution is orthogonal with our work.

{\bf Mixed speed training:}
To enhance model's delay-awareness, we also employ mixed speed training to extend the temporal perception range of Transtreamer. The model will be ineffective on other temporal intervals if only a fixed relative indices $\mathbf{P^P}$ and $\mathbf{P^F}$ is sampled for training. Therefore, we adopt mixed speed training scheme that samples $\mathbf{P^P}$ from $[-24, -1]$ and $\mathbf{P^F}$ from $[1, 16]$. The loss for each predicted future object is given equal weights.

\begin{table}
    \small
    \centering
    \tabcolsep=1pt\relax
    \caption{\small Main result of sAP comparison with real-time and non real-time SOTA detectors on the Argoverse-HD dataset. Best mAP scores are indicated in bold.}
    \label{tab: main}
    \begin{tabular}{lcccccc}
    \toprule
    Methods                           & $sAP$  & $sAP_{50}$ & $sAP_{75}$ & $sAP_S$ & $sAP_M$ & $sAP_L$ \\
    \midrule
    Non Real-time Methods & & & & & & \\
    \midrule
    Streamer (S=900)                  & 18.2 & 35.3  & 16.8  & 4.7  & 14.4 & 34.6 \\
    Streamer (S=600)                  & 20.4 & 35.6  & 20.8  & 3.6  & 18.0 & 47.2 \\
    Streamer + AdaS                   & 13.8 & 23.4  & 14.2  & 0.2  & 9.0  & 39.9 \\
    Adaptive Streamer                 & 21.3 & 37.3  & 21.1  & 4.4  & 18.7 & 47.1 \\
    YOLOX-S                           & 25.8 & 47.0  & 24.3  & 8.8  & 25.7 & 44.5 \\
    YOLOX-M                           & 29.4 & 51.6  & 28.1  & 10.3 & 29.9 & 50.4 \\
    YOLOX-L                           & 32.5 & 55.9  & 31.2  & 12.0 & 31.3 & 57.1 \\
    \midrule
    Real-time Methods & & & & & & \\
    \midrule
    StreamYOLO-S                      & 28.8 & 50.3  & 27.6  & 9.7  & 30.7 & 53.1 \\
    StreamYOLO-M                      & 32.9 & 54.0  & 32.5  & 12.4 & 34.8 & 58.1 \\
    StreamYOLO-L                      & 36.1 & 57.6  & 35.6  & 13.8 & 37.1 & 63.3 \\
    DADE-L                            & 36.7 & 63.9  & 36.9  & 14.6 & 57.9 & 37.3 \\
    LongShortNet-S                    & 29.8 & 50.4  & 29.5  & 11.0 & 30.6 & 52.8 \\
    LongShortNet-M                    & 34.1 & 54.8  & 34.6  & 13.3 & 35.3 & 58.1 \\
    LongShortNet-L                    & 37.1 & 57.8  & 37.7  & 15.2 & 37.3 & 63.8 \\
    DAMO-StreamNet-S                  & 31.8 & 52.3  & 31.0  & 11.4 & 32.9 & 58.7 \\
    DAMO-StreamNet-M                  & 35.7 & 56.7  & 35.9  & 14.5 & 36.3 & 63.3 \\
    DAMO-StreamNet-L                  & 37.8 & 59.1  & 38.6  & 16.1 & 39.0 & 64.6 \\
    Transtreaming-S(Ours)                    & 32.5 & 53.9  & 32.2  & 10.9 & 33.8 & 60.9 \\
    Transtreaming-M(Ours)                    & 36.0 & 57.0  & 36.6  & 14.2 & 36.9 & 63.6 \\
    Transtreaming-L(Ours)                    & \textbf{38.2} & \textbf{59.7}  & \textbf{39.0} & \textbf{16.3} & \textbf{40.0} & \textbf{65.8} \\
    \bottomrule
    \end{tabular}
\end{table}

\section{Experiment}
\subsection{Implementation Details}
{\bf Dataset:}
In our experiment, we trained and tested our method on Argoverse-HD, a common urban driving dataset composed of the front camera video sequence and bounding-box annotations for common road objects (e.g. cars, pedestrians, traffic lights). This dataset contains high-frequency annotation of 30 FPS that simulates real-world situations, which is suitable for streaming evaluation. We believe other datasets (e.g. nuScenes, Waymo) are not suitable for streaming evaluation as they are annotated at a lower frequency. We follow the train and validation split as in \cite{DAMO-StreamNet}.

{\bf Model:}
The base backbone of our proposed model is pretrained on the COCO dataset, consistent with the approach of \cite{DAMO-StreamNet}. Other parameters are initialized following Lecun weight initialization. The model is then fine-tuned on the Argoverse-HD dataset for 8 epochs using a single Nvidia RTX4080 GPU with a batch size of 4 and half-resolution input ($600 \times 960$). To ensure a fair comparison with other methods, we provide 3 configurations of Transtreaming as Transtreaming-S (small), Transtreaming-M (medium) and Transtreaming-L (large) that differs in the number of parameters. This is in accordance with previous methods' approaches \cite{DAMO-StreamNet, Longshortnet}.

\begin{table} 
    \small
    \centering
    \tabcolsep=1pt\relax
    \caption{\small sAP comparison for Computational Delay Adaptation with different real-world devices. * means the model is trained using mixed speed training technique. Best sAP scores are indicated in bold.}
    \label{tab: compute}
    \begin{tabular}{llcccc}
    \toprule
    Devices                        & Methods        & $sAP_2$ & $sAP_4$ & $sAP_8$ & $sAP_{16}$ \\
    \midrule
    \multirow{3}{*}{v100 cluster}  & LongShortNet-S         & 25.5 & 21.6 & 16.7 & 11.0  \\
                                   & DAMO-StreamNet-S        & 25.0 & 19.9 & 14.2 & 9.4   \\
                                   & Transtreaming-S*(Ours) & \textbf{26.4} & \textbf{22.4} & \textbf{16.8} & \textbf{11.8}  \\
    \midrule
    \multirow{3}{*}{4080 server}   & LongShortNet-S         & 24.5 & 20.3 & 14.8 & 9.8   \\
                                   & DAMO-StreamNet-S        & 23.9 & 18.5 & 12.7 & 8.8   \\
                                   & Transtreaming-S*(Ours) & \textbf{25.3} & \textbf{20.9} & \textbf{15.5} & \textbf{10.8}  \\
    \midrule
    \multirow{3}{*}{3090 server}   & LongShortNet-S         & 24.0 & 20.0 & 14.4 & 9.5   \\
                                   & DAMO-StreamNet-S        & 23.7 & 18.2 & 12.6 & 8.8   \\
                                   & Transtreaming-S*(Ours) & \textbf{24.9} & \textbf{20.8} & \textbf{15.1} & \textbf{10.4}  \\
    \midrule
    \multirow{3}{*}{2080Ti server} & LongShortNet-S         & 23.0 & 18.0 & 12.7 & 8.6   \\
                                   & DAMO-StreamNet-S        & 21.9 & 16.4 & 11.3 & 8.0   \\
                                   & Transtreaming-S*(Ours) & \textbf{23.5} & \textbf{19.0} & \textbf{13.2} & \textbf{9.1}   \\
    \bottomrule
    \end{tabular}
    
\end{table}

{\bf Main Metric:}
We follow the streaming evaluation methods proposed by \cite{Towards_Streaming_Perception} as the main test metric. The streaming Average Precision (sAP) is used to evaluated the performance of the whole pipeline under a simulated real-time situation. The sAP metric compares the output of model with ground-truth at output timestep, instead of input timestep as used in offline evaluations. sAP use the same calculation formula as mean Average Precision (mAP), computing the average precision scores of matched objects with IOU thresholds ranging from 0.5 to 0.95. The sAP scores for small, medium, and large objects (denoted as $sAP_S, sAP_M, sAP_L$ respectively) are also reported. To ensure a fair comparison, we did not employ mixed speed training in our model, in alignment with the training scheme of baseline methods.

{\bf Computation Delay Adaptation Metric:}
To simulate streaming perception on devices with diverse computation capabilities, we test the framework on 1 online GPU computing cluster (denoted as v100 cluster) and 3 real-world servers (denoted as 4080 server, 3090 serve and 2080Ti server) with different hardware specifications. The detailed information about these devices are listed in supplemental materials. Additionally, we adopt a delay factor $d$ to simulate high-delay situations, where all computation delays are multiplied by $d$. We assign $d \in \{2, 4, 8, 16\}$ and denote corresponding sAP value as $sAP_d$.

{\bf Vehicle Acceleration Adaptation Metric:}
To offer a more comprehensive comparison, we also evaluate the models (w/o strategy algorithm) under an offline setting using mean Average Precision (mAP). This approach ignores the impact of computational delay and only simulate different amplitude of vehicle accelerations, testing the model's ability to handle various objects displacement between adjacent frames. Different from common offline evaluation, the model is given past frames $\{\mathbf{I}_i\}$ and evaluated against future objects $\{\mathbf{O}_j\}, j \in \{ 2, 4, 8, 16\}$. The resulting mAP scores are denoted as $mAP_2, mAP_4, mAP_8, mAP_{16}$, respectively.

\subsection{Quantitative Results} 

{\bf Overall Performance Comparison:}
As the main result, our framework is evaluated against SOTA methods to demonstrate its strengths. Our proposed method achieves 38.2\% in sAP, surpassing current SOTA method by 0.4\%. Under S and M configurations, our pipeline also achieves the first place of most metrics, comparing to \cite{StreamYOLO}, \cite{Longshortnet} and \cite{DAMO-StreamNet}. The results clearly demonstrates the power of Transtreamer. Note that $sAP_L$ of our models is high, indicating that our proposed Transtreamer has recognized the temporal movement of large objects.

{\bf sAP Comparison for Computational Delay Adaptation:}
To demonstrate the robustness of the Transtreaming framework, we employ the Computation Delay Adaptation Metric and evaluate both our framework and baseline models across 4 devices with 4 different delay factor $d$. As illustrated in Table \ref{tab: compute}, our method surpasses current SOTA DAMO-StreamNet \cite{DAMO-StreamNet} by 1.1\%-2.8\% sAP on numerous real-world devices, demonstrating its strength on computation-bound environments. Notably, in high-latency scenarios, LongShortNet surpasses DAMO-StreamNet despite using a lighter backbone, indicating that large models do not necessarily scale effectively with increased latency. The result indicates that our method maintains its performance superiority even on devices with low computing power.

\begin{table}
    \small
    \centering
    \tabcolsep=1pt\relax
    \caption{\small Vehicle Acceleration Adaptation Comparison with different simulated speed variation. Best mAP scores are indicated in bold.}
    \label{tab: speed}
    \begin{tabular}{llcccc}
    \toprule
    Model sizes        & Methods                & $mAP_2$ & $mAP_4$ & $mAP_8$ & $mAP_{16}$ \\
    \midrule
    \multirow{4}{*}{S} & LongShortNet-S         & 26.4          & 20.6          & 13.4          & 8.9           \\
                       & DAMO-StreamNet-S       & 28.9          & 22.0          & 14.6          & 9.4           \\
                       & Transtreaming-S(Ours)  & \textbf{29.5} & 23.1          & 15.2          & 9.6           \\
                       & Transtreaming-S*(Ours) & 28.9          & \textbf{24.5} & \textbf{17.9} & \textbf{12.3} \\
    \midrule
    \multirow{4}{*}{M} & LongShortNet-M         & 30.7          & 23.8          & 15.8          & 9.7           \\
                       & DAMO-StreamNet-M       & 31.7          & 24.4          & 16.0          & 10.2          \\
                       & Transtreaming-M(Ours)  & \textbf{33.1} & 26.5          & 17.9          & 11.2          \\
                       & Transtreaming-M*(Ours) & 32.3          & \textbf{27.3} & \textbf{19.6} & \textbf{13.5} \\
    \midrule
    \multirow{4}{*}{L} & LongShortNet-L         & 33.2          & 25.8          & 17.2          & 10.7          \\
                       & DAMO-StreamNet-L       & 33.8          & 26.1          & 17.1          & 10.8          \\
                       & Transtreaming-L(Ours)  & \textbf{34.7} & 27.1          & 17.9          & 11.3          \\
                       & Transtreaming-L*(Ours) & 33.9          & \textbf{28.4} & \textbf{20.4} & \textbf{14.0} \\
    \bottomrule
    \end{tabular}
\end{table}

{\bf sAP Comparison for Vehicle Acceleration Adaptation:}
We also compared our framework on simulated circumstances with drastic vehicle accelerations. We evaluate our framework both with and w/o mixed speed training, as to investigate the influence of the strategy. As is shown in Table \ref{tab: speed}, our proposed framework achieves a absolute improvement of 0.6\%, 1.1\%, 0.6\%, 0.2\% under $mAP_2$, $mAP_4$, $mAP_8$ and $mAP_{16}$ respectively w/o mixed speed training compared to DAMO-StreamNet-S. Interestingly, though our models trained with mixed speed training only have a little advantage over other baseline models under $mAP_2$, a maximum 3.2\% mAP gain is observed on larger temporal intervals. The experimental results demonstrates the temporal-flexibility of our framework.

\subsection{Ablation study}
The results of the ablation study are listed in Table \ref{tab: ablation}. We verify the effectiveness of four proposed components: RTPE, TAT, Planner and Buffer. For RTPE, we remove it from our model and observed a 0.4\% decrease of test sAP. We also utilizes sinusoidal positional encoding and learnable positional encoding as substitutions, but both are inferior to our current approach. For TAT, we compare its performance with the neck used by \cite{Longshortnet, DAMO-StreamNet} and found a 0.9\% increase. As for Planner and Buffer which only operates under streaming settings, we examine their ability under $sAP_1$. We remove them from our framework and observed 0.4\% and an astonishing 25.8\% decrease on the metric. Since the detection model usually compute the features of 4 past frames, it is unrealistic to recompute 3 more frames' feature without huge sAP sacrifice.

\begin{table}
    \small
    \centering
    \tabcolsep=8pt
    \caption{\small Ablation study on Transtreaming-S. All the four components presented in this table are helpful in our framework. Since Planner and Buffer only functions under streaming tests, we split the Ablation study into two parts evaluated by mAP and sAP. Note the Buffer module is crucial in streaming perception settings by avoiding huge computation cost, both for Transtreaming and baseline methods}
    \label{tab: ablation}
    \centering
    \begin{tabular}{llccc}
    \toprule
    Planner & Buffer & $sAP$  & $sAP_{50}$ & $sAP_{75}$ \\
    \midrule
    $\times$       & $\times$      & 3.9  & 5.8     & 4.1 \\
    $\times$       & \checkmark      & 29.7 & 51.3    & 28.8 \\
    \checkmark       & \checkmark      & 30.1 & 51.7    & 29.4 \\
    \midrule
    RTPE & TAT & $sAP$  & $sAP_{50}$ & $sAP_{75}$ \\
    \midrule
    $\times$    & $\times$  & 29.0 & 50.1    & 29.1    \\
    $\times$    & \checkmark   & 29.9 & 51.7    & 29      \\
    \checkmark    & \checkmark   & 30.8 & 52.3    & 30.2  \\
    \bottomrule
    \end{tabular}
\end{table}

\section{Discussion}

{\bf Conclusion:}
We introduce Transtreaming, a novel streaming perception framework that utilizes temporal cues to produce multiple prediction that aligns with real-world time, effectively producing real-time detection results. Transtreaming is the pioneer framework in streaming perception that make use of temporal proposals and multi-frame output. Inspired from cross-attention, we design the Transtreamer detection model that handles input and output with dynamic temporal range. Adaptive Strategy is also proposed to provide buffer techniques and temporal cues to Transtreamer. Our framework not only outperforms current SOTA methods under ordinary streaming perception settings, but also surpasses SOTA methods by a large margin under high/dynamic computational delay and drastic vehicle acceleration environments.

{\bf Limitation:}
Despite the demonstrated strengths, Transtreaming still has its limitations. First, thought proposals $\mathbf{P^P}$ and $\mathbf{P^F}$ are produced to provide temporal information, Transtreamer only utilizes it in the computation of attention weights. A stronger integration could be achieved if it directly interacts with input tokens. Second, our Computation Delay Adaptation Metric will sometimes not reflect the true ability of the model due to temporal aliasing effect. In other words, the model will perform bad if computational time slightly misaligns with multiples of frame rate. Therefore, we should sample more values of $d$ to concretely evaluate the model. We leave these limitations for future work.

\clearpage

\bibliography{egbib}

\begin{thebibliography}{26}
\providecommand{\natexlab}[1]{#1}

\bibitem[{Bertasius, Torresani, and Shi(2018)}]{DCN}
Bertasius, G.; Torresani, L.; and Shi, J. 2018.
\newblock Object Detection in Video with Spatiotemporal Sampling Networks.
\newblock In \emph{Proceedings of the European Conference on Computer Vision
  (ECCV)}.

\bibitem[{Deng et~al.(2019)Deng, Pan, Yao, Zhou, Li, and Mei}]{RDN}
Deng, J.; Pan, Y.; Yao, T.; Zhou, W.; Li, H.; and Mei, T. 2019.
\newblock Relation Distillation Networks for Video Object Detection.
\newblock In \emph{Proceedings of the IEEE/CVF International Conference on
  Computer Vision (ICCV)}.

\bibitem[{Feichtenhofer, Pinz, and Zisserman(2017)}]{D&T}
Feichtenhofer, C.; Pinz, A.; and Zisserman, A. 2017.
\newblock Detect to Track and Track to Detect.
\newblock In \emph{Proceedings of the IEEE International Conference on Computer
  Vision (ICCV)}.

\bibitem[{Ge et~al.(2021)Ge, Liu, Wang, Li, and Sun}]{yolox}
Ge, Z.; Liu, S.; Wang, F.; Li, Z.; and Sun, J. 2021.
\newblock YOLOX: Exceeding YOLO Series in 2021.
\newblock \emph{arXiv preprint arXiv:2107.08430}.

\bibitem[{Girshick(2015)}]{FastRCNN}
Girshick, R. 2015.
\newblock Fast R-CNN.
\newblock In \emph{2015 IEEE International Conference on Computer Vision
  (ICCV)}, 1440--1448.

\bibitem[{Girshick et~al.(2014)Girshick, Donahue, Darrell, and
  Malik}]{girshick_rich_2014}
Girshick, R.; Donahue, J.; Darrell, T.; and Malik, J. 2014.
\newblock Rich feature hierarchies for accurate object detection and semantic
  segmentation.
\newblock ArXiv:1311.2524 [cs].

\bibitem[{Han et~al.(2020)Han, Wang, Chang, and Qiao}]{HVRNet}
Han, M.; Wang, Y.; Chang, X.; and Qiao, Y. 2020.
\newblock Mining Inter-Video Proposal Relations for Video Object Detection.
\newblock In Vedaldi, A.; Bischof, H.; Brox, T.; and Frahm, J.-M., eds.,
  \emph{Computer Vision -- ECCV 2020}, 431--446. Cham: Springer International
  Publishing.
\newblock ISBN 978-3-030-58589-1.

\bibitem[{He et~al.(2023)He, Cheng, Li, Xiang, Chen, Luo, Geng, and
  Xie}]{DAMO-StreamNet}
He, J.-Y.; Cheng, Z.-Q.; Li, C.; Xiang, W.; Chen, B.; Luo, B.; Geng, Y.; and
  Xie, X. 2023.
\newblock DAMO-StreamNet: Optimizing Streaming Perception in Autonomous
  Driving.
\newblock \emph{arXiv preprint arXiv:2303.17144}.

\bibitem[{Huang and Chen(2023)}]{MTD}
Huang, Y.; and Chen, N. 2023.
\newblock MTD: Multi-Timestep Detector for Delayed Streaming Perception.
\newblock \emph{arXiv preprint arXiv:2309.06742}.

\bibitem[{Jo et~al.(2022)Jo, Lee, Baik, Lee, Choi, and Park}]{DaDe}
Jo, W.; Lee, K.; Baik, J.; Lee, S.; Choi, D.; and Park, H. 2022.
\newblock DaDe: Delay-adoptive Detector for Streaming Perception.
\newblock \emph{arXiv preprint arXiv:2212.11558}.

\bibitem[{Li et~al.(2023)Li, Cheng, He, Li, Luo, Chen, Geng, Lan, and
  Xie}]{Longshortnet}
Li, C.; Cheng, Z.-Q.; He, J.-Y.; Li, P.; Luo, B.; Chen, H.; Geng, Y.; Lan,
  J.-P.; and Xie, X. 2023.
\newblock Longshortnet: Exploring temporal and semantic features fusion in
  streaming perception.
\newblock In \emph{ICASSP 2023-2023 IEEE International Conference on Acoustics,
  Speech and Signal Processing (ICASSP)}, 1--5. IEEE.

\bibitem[{Li, Wang, and Ramanan(2020)}]{Towards_Streaming_Perception}
Li, M.; Wang, Y.-X.; and Ramanan, D. 2020.
\newblock Towards streaming perception.
\newblock In \emph{Computer Vision--ECCV 2020: 16th European Conference,
  Glasgow, UK, August 23--28, 2020, Proceedings, Part II 16}, 473--488.
  Springer.

\bibitem[{Liu and Zhu(2018)}]{LSTM}
Liu, M.; and Zhu, M. 2018.
\newblock Mobile Video Object Detection With Temporally-Aware Feature Maps.
\newblock In \emph{Proceedings of the IEEE Conference on Computer Vision and
  Pattern Recognition (CVPR)}.

\bibitem[{Liu et~al.(2016)Liu, Anguelov, Erhan, Szegedy, Reed, Fu, and
  Berg}]{SSD}
Liu, W.; Anguelov, D.; Erhan, D.; Szegedy, C.; Reed, S.; Fu, C.-Y.; and Berg,
  A. 2016.
\newblock SSD: Single Shot MultiBox Detector.
\newblock In \emph{European Conference on Computer Vision}, volume 9905,
  21--37.
\newblock ISBN 978-3-319-46447-3.

\bibitem[{Liu et~al.(2021)Liu, Lin, Cao, Hu, Wei, Zhang, Lin, and
  Guo}]{liu2021swin}
Liu, Z.; Lin, Y.; Cao, Y.; Hu, H.; Wei, Y.; Zhang, Z.; Lin, S.; and Guo, B.
  2021.
\newblock Swin transformer: Hierarchical vision transformer using shifted
  windows.
\newblock In \emph{Proceedings of the IEEE/CVF international conference on
  computer vision}, 10012--10022.

\bibitem[{Parmar et~al.(2018)Parmar, Vaswani, Uszkoreit, Kaiser, Shazeer, Ku,
  and Tran}]{parmar_image_2018}
Parmar, N.; Vaswani, A.; Uszkoreit, J.; Kaiser, L.; Shazeer, N.; Ku, A.; and
  Tran, D. 2018.
\newblock Image {Transformer}.
\newblock ArXiv:1802.05751 [cs].

\bibitem[{Redmon et~al.(2016)Redmon, Divvala, Girshick, and Farhadi}]{YOLO}
Redmon, J.; Divvala, S.; Girshick, R.; and Farhadi, A. 2016.
\newblock You Only Look Once: Unified, Real-Time Object Detection.
\newblock In \emph{2016 IEEE Conference on Computer Vision and Pattern
  Recognition (CVPR)}, 779--788.

\bibitem[{Ren et~al.(2015)Ren, He, Girshick, and Sun}]{FasterRCNN}
Ren, S.; He, K.; Girshick, R.; and Sun, J. 2015.
\newblock Faster {R}-{CNN}: {Towards} {Real}-{Time} {Object} {Detection} with
  {Region} {Proposal} {Networks}.
\newblock In \emph{Advances in {Neural} {Information} {Processing} {Systems}},
  volume~28. Curran Associates, Inc.

\bibitem[{Shvets, Liu, and Berg(2019)}]{LLTR}
Shvets, M.; Liu, W.; and Berg, A.~C. 2019.
\newblock Leveraging Long-Range Temporal Relationships Between Proposals for
  Video Object Detection.
\newblock In \emph{Proceedings of the IEEE/CVF International Conference on
  Computer Vision (ICCV)}.

\bibitem[{Wang et~al.(2018)Wang, Zhou, Yan, and Deng}]{MANet}
Wang, S.; Zhou, Y.; Yan, J.; and Deng, Z. 2018.
\newblock Fully Motion-Aware Network for Video Object Detection.
\newblock In \emph{Proceedings of the European Conference on Computer Vision
  (ECCV)}.

\bibitem[{Wu et~al.(2019)Wu, Chen, Wang, and Zhang}]{SELSA}
Wu, H.; Chen, Y.; Wang, N.; and Zhang, Z. 2019.
\newblock Sequence Level Semantics Aggregation for Video Object Detection.
\newblock In \emph{Proceedings of the IEEE/CVF International Conference on
  Computer Vision (ICCV)}.

\bibitem[{Xiao and Lee(2018)}]{RNN}
Xiao, F.; and Lee, Y.~J. 2018.
\newblock Video Object Detection with an Aligned Spatial-Temporal Memory.
\newblock In \emph{Proceedings of the European Conference on Computer Vision
  (ECCV)}.

\bibitem[{Yang et~al.(2022)Yang, Liu, Li, Li, and Sun}]{StreamYOLO}
Yang, J.; Liu, S.; Li, Z.; Li, X.; and Sun, J. 2022.
\newblock Real-time object detection for streaming perception.
\newblock In \emph{Proceedings of the IEEE/CVF conference on computer vision
  and pattern recognition}, 5385--5395.

\bibitem[{Zhu et~al.(2018)Zhu, Dai, Yuan, and Wei}]{THP}
Zhu, X.; Dai, J.; Yuan, L.; and Wei, Y. 2018.
\newblock Towards High Performance Video Object Detection.
\newblock In \emph{Proceedings of the IEEE Conference on Computer Vision and
  Pattern Recognition (CVPR)}.

\bibitem[{Zhu et~al.(2017{\natexlab{a}})Zhu, Wang, Dai, Yuan, and Wei}]{FGFA}
Zhu, X.; Wang, Y.; Dai, J.; Yuan, L.; and Wei, Y. 2017{\natexlab{a}}.
\newblock Flow-Guided Feature Aggregation for Video Object Detection.
\newblock In \emph{Proceedings of the IEEE International Conference on Computer
  Vision (ICCV)}.

\bibitem[{Zhu et~al.(2017{\natexlab{b}})Zhu, Xiong, Dai, Yuan, and Wei}]{DFF}
Zhu, X.; Xiong, Y.; Dai, J.; Yuan, L.; and Wei, Y. 2017{\natexlab{b}}.
\newblock Deep Feature Flow for Video Recognition.
\newblock In \emph{Proceedings of the IEEE Conference on Computer Vision and
  Pattern Recognition (CVPR)}.

\end{thebibliography}

\end{document}